\documentclass[11pt,a4paper]{article}
\usepackage[hyperref]{emnlp2020}
\usepackage{times}
\usepackage{natbib}
\usepackage{latexsym}

\usepackage{microtype}
\usepackage{graphicx}
\usepackage{blindtext}

\aclfinalcopy 

\usepackage{booktabs}

\newcommand{\parheader}[1]{{\smallskip \noindent \bf #1.}}

\title{Howl: A Deployed, Open-Source Wake Word Detection System}

\author{Raphael Tang,$^1$\thanks{\hspace{2mm}Equal contribution. Order decided by coin flip.} \hspace{0.125cm}Jaejun Lee,$^{1*}$ Afsaneh Razi,$^2$ Julia Cambre,$^2$\\ {\bf Ian Bicking,$^2$ Jofish Kaye,$^2$ \and Jimmy Lin$^1$}\\
$^1$David R. Cheriton School of Computer Science, University of Waterloo\\
$^2$Mozilla}

\date{}

\begin{document}
\maketitle
\begin{abstract}
We describe Howl, an open-source wake word detection toolkit with native support for open speech datasets, like Mozilla Common Voice and Google Speech Commands.
We report benchmark results on Speech Commands and our own freely available wake word detection dataset, built from MCV.
We operationalize our system for Firefox Voice, a plugin enabling speech interactivity for the Firefox web browser.
Howl represents, to the best of our knowledge, the first fully productionized yet open-source wake word detection toolkit with a web browser deployment target.
Our codebase is at \url{https://github.com/castorini/howl}.

\end{abstract}


\section{Introduction}
Wake word detection is the task of recognizing an utterance for activating a speech assistant, such as ``Hey, Alexa'' for the Amazon Echo.
Given that such systems are meant to support full automatic speech recognition, the task seems simple; however, it introduces a different set of challenges because these systems have to be always listening, computationally efficient, and, most of all, privacy respecting.
Therefore, the community treats it as a separate line of work, with most recent advancements driven predominantly by neural networks~\cite{sainath2015convolutional, tang2018deep}.

Unfortunately, most existing toolkits are closed source and often specific to a target platform.
Such design choices restrict the flexibility of the application and add unnecessary maintenance as the number of target domains increases.
We argue that using JavaScript is a solution: unlike many languages and their runtimes, the JavaScript engine powers a wide range of modern user-facing applications ranging from mobile  to desktop ones.

To this end, we have previously developed Honkling, a JavaScript-based keyword spotting system~\cite{lee2019honkling}.
Leveraging one of the lightest models available for the task from \citet{tang2018deep}, Honkling efficiently detects the target commands with high precision. 
However, we notice that Honkling is still quite far from being a stable wake word detection system.
This gap mainly arises from the model being trained as a speech commands classifier, instead of a wake word detector; its high false alarm rate results from the limited number of negative samples in the training dataset~\cite{warden2018speech}.

In this paper, to make a greater practical impact, we close this gap in the Honkling ecosystem and present Howl, an open-source wake word detection toolkit with support for open datasets such as Mozilla Common Voice (MCV; \citealp{ardila2019common}) and the Google Speech Commands dataset~\cite{warden2018speech}.
Our new system is the first in-browser wake word system which powers a widely deployed industrial application, Firefox Voice.
By processing the audio in the browser and being completely open source, including the datasets and models, Howl is a privacy-respecting, non-eavesdropping system which users can trust.
Having a false reject rate of 10\% at 4 false alarms per hour of speech, Howl has enabled Firefox Voice to provide a completely hands-free experience to over 8,000 users in the nine days since its launch.

\begin{figure*}[!t]
  \centering
  \includegraphics[scale=0.55]{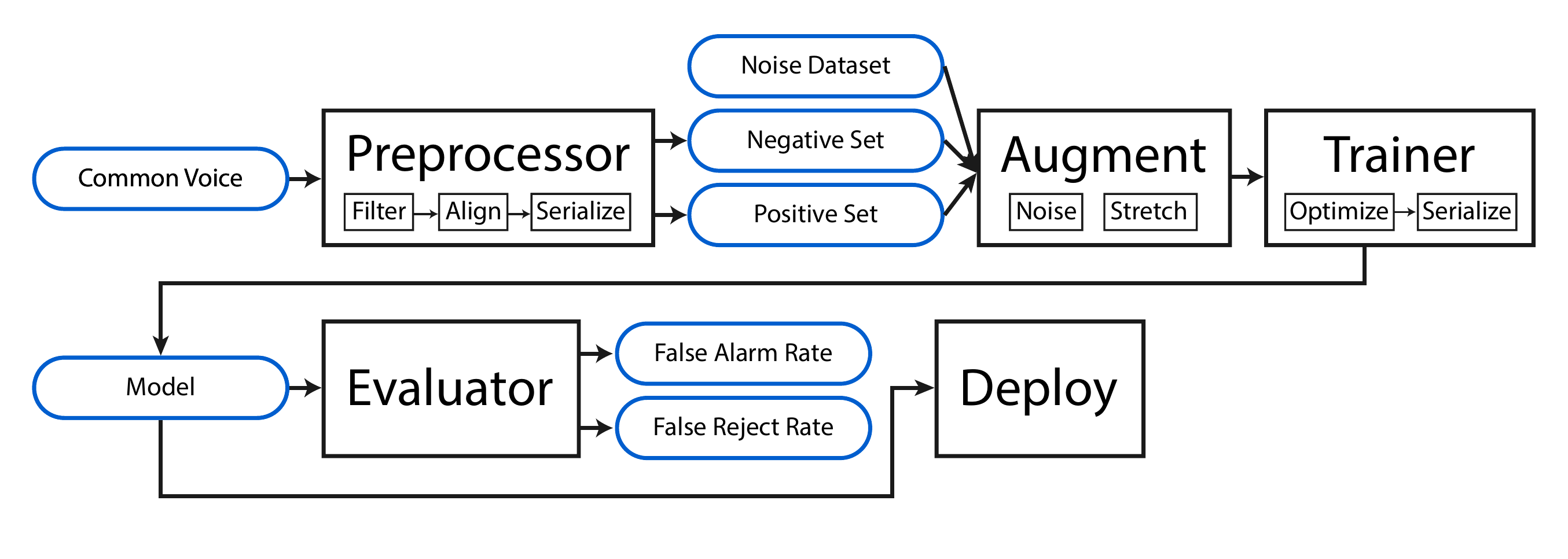}
  \caption{An illustration of the pipeline and its control flow. First, we preprocess Common Voice by filtering for the wake word vocabulary, aligning the speech, and saving the negative and positives sets to disk. Next, we introduce a noise dataset and augment the data on the fly at training time. Finally, we evaluate the optimized model and, if the results are satisfactory, export it for deployment.}
  \label{fig:pipeline}
\end{figure*}

\section{Background and Related Work}
Other than privately owned wake word detection systems, Porcupine and Snowboy are the most well-known ecosystems that provide an open-source modeling toolkit, some data, and deployment capabilities.
However, these ecosystems are still closed at heart; they keep their data, models, or deployment proprietary.
As far as open-source ecosystems go, Precise\footnote{\url{https://github.com/MycroftAI/mycroft-precise}} represents a step in the right direction, but its datasets are limited, and its deployment target is the Raspberry Pi.



We further make the distinction from speech commands classification toolkits, such as Honk~\cite{tang2017honk}.
These frameworks focus on classifying fixed-length audio as one of a few dozen keywords, with no evaluation on a sizable negative set, as required in wake word detection.
While these trained models may be used in detection applications, they are not rigorously tested for such.


\section{System}

We present a high-level description of our toolkit and its goals.
For specific details, we refer users to the repository, as linked in the abstract.

\subsection{Requirements}
Howl is written in Python 3.7+, with the notable dependencies being PyTorch for model training, Librosa~\cite{mcfee2015librosa} for audio preprocessing, and Montreal Forced Aligner~(MFA; \citealp{mcauliffe2017montreal}) for speech data alignment.
We license Howl under the Mozilla Public License v2, a file-level copyleft free license.
For speedy model training, we recommend a CUDA-enabled graphics card with at least 4GB of VRAM; we used an Nvidia Titan RTX in all of our experiments.
The rest of the computer can be built with, say, 16GB of RAM and a mid-range desktop CPU.
For resource-restricted users, we suggest exploring Google Colab\footnote{\url{https://colab.research.google.com/}} and other cloud-based solutions.

\subsection{Components and Pipeline}
Howl consists of the three following major components: audio preprocessing, data augmentation, and model training and evaluation.
These components form a pipeline, in the written order, for producing deployable models from raw audio data.

\parheader{Preprocessing}
A wake word dataset must first be preprocessed from an annotated data source, which is defined as a collection of audio--transcription pairs, with predefined training, development, and test splits.
Since Howl is a frame-level keyword spotting system, it relies on a forced aligner to provide word- or phone-based alignment.
We choose MFA for its popularity and free license, and hence Howl structures the processed datasets to interface well with MFA.

Another preprocessing task is to parse the global configuration settings for the framework.
Such settings include the learning rate, the dataset path, and model-specific hyperparameters.
We read in most of these settings as environment variables, which enable easy shell scripting.

\parheader{Augmentation}
For improved robustness and better quality, we implement a set of popular augmentation routines: time stretching, time shifting, synthetic noise addition, recorded noise mixing, SpecAugment (no time warping; ~\citealp{park2019specaugment}), and vocal tract length perturbation~\cite{jaitly2013vocal}.
These are readily extensible, so practitioners may easily add new augmentation modules.

\parheader{Training and evaluation}
Howl provides several off-the-shelf neural models, as well as training and evaluation routines using PyTorch for computing the loss gradient and the task-specific metrics, such as the false alarm rate and reject rate.
These routines are also responsible for serializing the model and exporting it to our browserside deployment.

\parheader{Pipeline}
Given these components, our pipeline, visually presented in Figure~\ref{fig:pipeline}, is as follows: First, users produce a wake word detection dataset, either manually or from a data source like Common Voice and Google Speech Commands, setting the appropriate environment variables.
This can be quickly accomplished using Common Voice, whose ample breadth and coverage of popular English words allow for a wide selection of custom wake words; for example, it has about a thousand occurrences of the word ``next.''
In addition to a positive subset containing the vocabulary and wake word, this dataset ideally contains a sizable negative set, which is necessary for more robust models and a more accurate evaluation of the false positive rate.

Next, users (optionally) select which augmentation modules to use, and they train a model with the provided hyperparameters on the selected dataset, which is first processed into log-Mel frames with zero mean and unit variance, as is standard.
This training process should take less than a few hours on a GPU-capable device for most use cases, including ours.
Finally, users may run the model in the included command line interface demo or deploy it to the browser using Honkling, our in-browser keyword spotting (KWS) system, if the model is supported~\cite{lee2019honkling}.

\subsection{Data and Models}
For the data sources, Howl works out of the box with MCV, a general speech corpus, and Speech Commands, a commands recognition dataset.
Users can quickly extend Howl to accept other speech corpuses such as LibriSpeech~\cite{panayotov2015librispeech}.
Howl also accepts any folder that contains audio files and interprets them as recorded noise for data augmentation, which covers noise datasets such as MUSAN~\cite{snyder2015musan} and Microsoft SNSD~\cite{reddy2019scalable}.

For modeling, Howl provides implementations of convolutional neural networks (CNNs) and recurrent neural networks (RNNs) for wake word detection.
These models are from the existing literature, such as residual CNNs~\cite{tang2018deep}, a modified listen--attend--spell (LAS) encoder~\cite{chan2015listen, park2019specaugment}, and MobileNetv2~\cite{sandler2018mobilenetv2}.
Most of the models are lightweight since the end application requires efficient inference, though some are parameter heavy to establish a rough upper bound on the quality, as far as parameters go.
Of particular focus is the lightweight \texttt{res8} model~\cite{tang2018deep}, which is directly exportable to Honkling, the in-browser KWS system.
For this reason, we choose it in our deployment to Firefox Voice.\footnote{\url{https://github.com/mozilla-extensions/firefox-voice}}


\begin{table}[t]
    \centering
    \setlength{\tabcolsep}{1pt}
    \begin{tabular}{l c c c}
        \toprule[1pt]
        Model & Dev/Test & \# Par. \\
        \midrule 
        EdgeSpeechNet~\cite{lin2018edgespeechnets} & --/96.8 & 107K \\
        \texttt{res8}~\cite{tang2018deep} & --/94.1 & 110K \\
        RNN~\cite{de2018neural} & --/95.6 & 202K \\
        DenseNet~\cite{zeng2019effective} & --/97.5 & 250K \\
        \midrule
        Our \texttt{res8} & {\bf 97.0}/\textbf{97.8} & 111K\\
        Our LSTM & 94.3/94.5 & 128K\\
        Our LAS encoder & 96.8/97.1 & 478K\\
        Our MobileNetv2 & 96.4/97.3 & 2.3M\\
        \bottomrule[1pt]
    \end{tabular}
    \caption{Model accuracy on Google Speech Commands. Bolded denotes the best and \# par. the number of parameters.}
    \label{table:gsc-results}
\end{table}

\section{Benchmark Results}
To verify the correctness of our implementation, we first train and evaluate our models on the Google Speech Commands dataset, for which there exists many known results.
Next, we curate a wake word detection datasets and report our resulting model quality.
Training details are in the repository.

\parheader{Commands recognition}
We report in Table~\ref{table:gsc-results} the results of the twelve-keyword recognition task from Speech Commands (v1), where we classify a one-second clip as one of ``yes,'' ``no,'' ``up,'' ``down,'' ``left,'' ``right,'' ``on,'' ``off,'' ``stop,'' ``go,'' unknown, or silence.
Our implementations are competitive with state of the art, with the \texttt{res8} model surprisingly achieving the highest accuracy of 97.8 on the test set, despite having fewer parameters.
Our other implemented models, the LSTM, LAS encoder, and MobileNetv2, compare favorably.

\parheader{Wake word detection}
\begin{figure}
    \centering
    \includegraphics[scale=0.78]{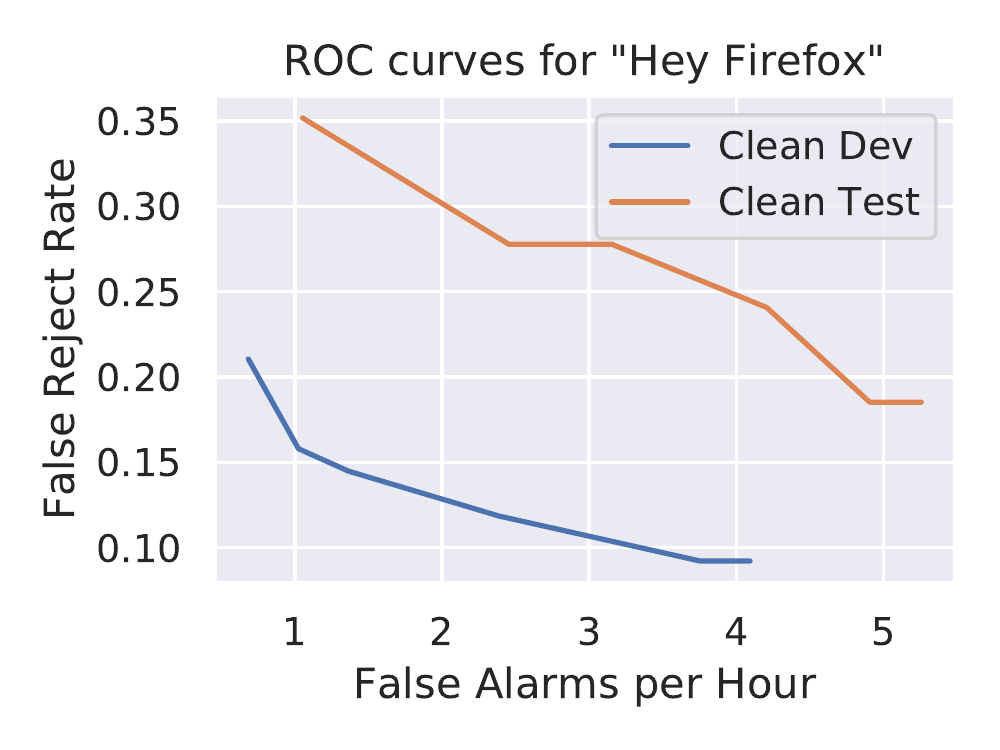}
    \caption{Receiver operating characteristic (ROC) curves for the wake word.}
    \label{fig:result-heyff}
\end{figure}
For wake word detection, we target ``hey, Firefox'' for waking up Firefox Voice.
From the single-word segment of MCV, we use 1,894 and 1,877 recordings of ``hey'' and ``Firefox,'' respectively; from the MCV general speech corpus, we select all 1,037 recordings containing ``hey,'' ``fire,'' or ``fox.''
We additionally collect 632 recordings of ``hey, Firefox'' from volunteers.
For the negative set, we use about 10\% of the entire MCV speech corpus.
We choose the training, dev, and test splits to be 80\%, 10\%, and 10\% of the resulting corpus, stratified by speaker IDs for the positive set.
For robustness to noise, we use portions of MUSAN and SNSD as the noise dataset.
We arrive at 31 hours of data for training and 3 hours each for dev and test.

For the model, we select \texttt{res8}~\cite{tang2018deep} for its high quality on Speech Commands and easy adaptability with our browser deployment target.
We follow the aforementioned pipeline to train it; details are not repeated, and hyperparameters can be found in the repository.

We present the resulting receiver operating characteristic curves in Figure~\ref{fig:result-heyff}, where different operating points result from different thresholds on the output probabilities.
Although it seems to lag commercial systems~\cite{sainath2015convolutional} by 10--20\% at the same number of false alarms per hour, those systems are trained with 5--20$\times$ more data.
Our negative set also likely contains more adversarial examples that misrepresent real-world usage, e.g., many utterances of ``Firefox,'' which are responsible for at least 90\% of the false positives.
Thus, combined with favorable though preliminary results from live testing the system ourselves, we comfortably choose the operating point at four false alarms per hour.
We finally note that the discrepancy between the dev and test curves is likely explained by differences in the data distribution, not hyperparameter fiddling, because there are only 76 and 54 clips in the positive dev and test sets, respectively.

\section{Browser Deployment}

To protect user security and privacy, wake word detection must be achieved with the user's resources only. 
This setting introduces various technical challenges, as the available resources are often limited and may not be accessible.
In the case of Firefox Voice, our target application, the platform is Firefox, where the major challenge is the limited support in machine learning frameworks. 

However, our previous line of work demonstrates the feasibility of in-browser wake word detection with Honkling~\cite{lee2019honkling}.
Our application is written purely in JavaScript and supports different models using TensorFlow.js.
During the process of integrating Honkling with Firefox Voice, the two main aspects we focus on are accuracy and efficiency.
We rewrite the audio processing logic of Honkling to match the new Python pipeline and optimize various preprocessing routines to substantially reduce the computational burden.

To measure the performance of our application, we refer to the built-in energy impact metric of Firefox, which reports the CPU consumption of each open tab.
To establish a reference, playing a YouTube video reports an average energy impact of 10, while a static Google search reports 0.1.
Fortunately, our wake word detection model yields an energy impact of only 3, which efficiently enables hands-free interaction for initiating the speech recognition engine.
Our wake word detection demo and browserside integration details can be found at \url{https://github.com/castorini/howl-deploy}.



\section{Conclusions and Future Work}

In this paper, we introduce Howl, the first in-browser wake word detection system which powers a widely deployed application, Firefox Voice.
Leveraging a continuously growing speech dataset, Howl enables a communal endeavour for building a privacy-respecting and non-eavesdropping wake word detection system.
To expand the scope of Howl, our future work includes embedded systems as deployment targets, where the computational resources are much more constrained, with some systems lacking even modern memory managers.

\bibliographystyle{acl_natbib}
\bibliography{emnlp2020}

\begin{thebibliography}{18}
\expandafter\ifx\csname natexlab\endcsname\relax\def\natexlab#1{#1}\fi

\bibitem[{de~Andrade et~al.(2018)de~Andrade, Leo, Viana, and
  Bernkopf}]{de2018neural}
Douglas~Coimbra de~Andrade, Sabato Leo, Martin Loesener Da~Silva Viana, and
  Christoph Bernkopf. 2018.
\newblock A neural attention model for speech command recognition.
\newblock \emph{arXiv:1808.08929}.

\bibitem[{Ardila et~al.(2019)Ardila, Branson, Davis, Henretty, Kohler, Meyer,
  Morais, Saunders, Tyers, and Weber}]{ardila2019common}
Rosana Ardila, Megan Branson, Kelly Davis, Michael Henretty, Michael Kohler,
  Josh Meyer, Reuben Morais, Lindsay Saunders, Francis~M. Tyers, and Gregor
  Weber. 2019.
\newblock Common voice: A massively-multilingual speech corpus.
\newblock \emph{arXiv:1912.06670}.

\bibitem[{Chan et~al.(2015)Chan, Jaitly, Le, and Vinyals}]{chan2015listen}
William Chan, Navdeep Jaitly, Quoc~V. Le, and Oriol Vinyals. 2015.
\newblock Listen, attend and spell.
\newblock \emph{arXiv:1508.01211}.

\bibitem[{Jaitly and Hinton(2013)}]{jaitly2013vocal}
Navdeep Jaitly and Geoffrey~E. Hinton. 2013.
\newblock Vocal tract length perturbation ({VTLP}) improves speech recognition.
\newblock In \emph{Proceedings of the ICML Workshop on Deep Learning for Audio,
  Speech and Language}.

\bibitem[{Lee et~al.(2019)Lee, Tang, and Lin}]{lee2019honkling}
Jaejun Lee, Raphael Tang, and Jimmy Lin. 2019.
\newblock Honkling: {I}n-browser personalization for ubiquitous keyword
  spotting.
\newblock In \emph{Proceedings of the 2019 Conference on Empirical Methods in
  Natural Language Processing}.

\bibitem[{Lin et~al.(2018)Lin, Chung, and Wong}]{lin2018edgespeechnets}
Zhong~Qiu Lin, Audrey~G. Chung, and Alexander Wong. 2018.
\newblock {EdgeSpeechNets}: Highly efficient deep neural networks for speech
  recognition on the edge.
\newblock \emph{arXiv:1810.08559}.

\bibitem[{McAuliffe et~al.(2017)McAuliffe, Socolof, Mihuc, Wagner, and
  Sonderegger}]{mcauliffe2017montreal}
Michael McAuliffe, Michaela Socolof, Sarah Mihuc, Michael Wagner, and Morgan
  Sonderegger. 2017.
\newblock Montreal forced aligner: Trainable text-speech alignment using
  {K}aldi.
\newblock In \emph{Proceedings of the Eighteenth Annual Conference of the
  International Speech Communication Association}.

\bibitem[{McFee et~al.(2015)McFee, Raffel, Liang, Ellis, McVicar, Battenberg,
  and Nieto}]{mcfee2015librosa}
Brian McFee, Colin Raffel, Dawen Liang, Daniel P.~W. Ellis, Matt McVicar, Eric
  Battenberg, and Oriol Nieto. 2015.
\newblock librosa: Audio and music signal analysis in {P}ython.
\newblock In \emph{Proceedings of the 14th {P}ython in Science Conference}.

\bibitem[{Panayotov et~al.(2015)Panayotov, Chen, Povey, and
  Khudanpur}]{panayotov2015librispeech}
Vassil Panayotov, Guoguo Chen, Daniel Povey, and Sanjeev Khudanpur. 2015.
\newblock Librispeech: an {ASR} corpus based on public domain audio books.
\newblock In \emph{Proceedings of the IEEE International Conference on
  Acoustics, Speech and Signal Processing}.

\bibitem[{Park et~al.(2019)Park, Chan, Zhang, Chiu, Zoph, Cubuk, and
  Le}]{park2019specaugment}
Daniel~S. Park, William Chan, Yu~Zhang, Chung-Cheng Chiu, Barret Zoph,
  Ekin~Dogus Cubuk, and Quoc~V. Le. 2019.
\newblock Spec{A}ugment: A simple augmentation method for automatic speech
  recognition.
\newblock In \emph{Proceedings of the Twentieth Annual Conference of the
  International Speech Communication Association}.

\bibitem[{Reddy et~al.(2019)Reddy, Beyrami, Pool, Cutler, Srinivasan, and
  Gehrke}]{reddy2019scalable}
Chandan K.~A. Reddy, Ebrahim Beyrami, Jamie Pool, Ross Cutler, Sriram
  Srinivasan, and Johannes Gehrke. 2019.
\newblock A scalable noisy speech dataset and online subjective test framework.
\newblock In \emph{Proceedings of the Twentieth Annual Conference of the
  International Speech Communication Association}.

\bibitem[{Sainath and Parada(2015)}]{sainath2015convolutional}
Tara~N. Sainath and Carolina Parada. 2015.
\newblock Convolutional neural networks for small-footprint keyword spotting.
\newblock In \emph{Proceedings of the Sixteenth Annual Conference of the
  International Speech Communication Association}.

\bibitem[{Sandler et~al.(2018)Sandler, Howard, Zhu, Zhmoginov, and
  Chen}]{sandler2018mobilenetv2}
Mark Sandler, Andrew Howard, Menglong Zhu, Andrey Zhmoginov, and Liang-Chieh
  Chen. 2018.
\newblock Mobile{N}etv2: Inverted residuals and linear bottlenecks.
\newblock In \emph{Proceedings of the IEEE Conference on Computer Vision and
  Pattern Recognition}.

\bibitem[{Snyder et~al.(2015)Snyder, Chen, and Povey}]{snyder2015musan}
David Snyder, Guoguo Chen, and Daniel Povey. 2015.
\newblock {MUSAN}: A music, speech, and noise corpus.
\newblock \emph{arXiv:1510.08484}.

\bibitem[{Tang and Lin(2017)}]{tang2017honk}
Raphael Tang and Jimmy Lin. 2017.
\newblock Honk: A {P}y{T}orch reimplementation of convolutional neural networks
  for keyword spotting.
\newblock \emph{arXiv:1710.06554}.

\bibitem[{Tang and Lin(2018)}]{tang2018deep}
Raphael Tang and Jimmy Lin. 2018.
\newblock Deep residual learning for small-footprint keyword spotting.
\newblock In \emph{Proceedings of the IEEE International Conference on
  Acoustics, Speech and Signal Processing}.

\bibitem[{Warden(2018)}]{warden2018speech}
Pete Warden. 2018.
\newblock Speech commands: A dataset for limited-vocabulary speech recognition.
\newblock \emph{arXiv:1804.03209}.

\bibitem[{Zeng and Xiao(2019)}]{zeng2019effective}
Mengjun Zeng and Nanfeng Xiao. 2019.
\newblock Effective combination of densenet and bilstm for keyword spotting.
\newblock \emph{IEEE Access}, 7:10767--10775.

\end{thebibliography}

\end{document}